\documentclass[10pt,journal,compsoc]{IEEEtran}
\usepackage{marginnote}

\usepackage[style=ieee]{biblatex} 
\usepackage{hyperref}
\ifCLASSINFOpdf
   \usepackage[pdftex]{graphicx}
    \graphicspath{ {./src/img/} }
\else
\fi
\usepackage{tabularx}
\newenvironment{conditions*}
  {\par\vspace{\abovedisplayskip}\noindent
  \tabularx{\columnwidth}{>{$}l<{$} @{${}={}$} >{\raggedright\arraybackslash}X}}
  {\endtabularx\par\vspace{\belowdisplayskip}}

\usepackage{amssymb}
\usepackage{mathtools}

\usepackage[noend]{algpseudocode}
\usepackage{algorithm}
\begin{document}
%
\title{EDLaaS: Fully Homomorphic Encryption Over Neural Network Graphs for Vision and Private Strawberry Yield Forecasting}
%
%
%
%

\author{George Onoufriou,
        Marc Hanheide,
        and~Georgios Leontidis
\IEEEcompsocitemizethanks{\IEEEcompsocthanksitem G. Onoufriou and M. Hanheide are both with the University of Lincoln.\protect\\
E-mail: GOnoufriou at lincoln dot ac dot uk
\IEEEcompsocthanksitem Georgios Leontidis is with with the University of Aberdeen.}
\thanks{Manuscript received \today; revised never.}}

%
%

\markboth{Journal of \LaTeX\ Class Files,~Vol.~14, No.~8, August~2015}%
{Shell \MakeLowercase{\textit{et al.}}: Bare Demo of IEEEtran.cls for Computer Society Journals}
%



\IEEEtitleabstractindextext{%
    \begin{abstract}
      We present automatically parameterised Fully Homomorphic Encryption (FHE) for encrypted neural network inference and exemplify our inference over FHE compatible neural networks with our own open-source framework and reproducible examples. We use the 4th generation Cheon, Kim, Kim and Song (CKKS) FHE scheme over fixed points provided by the Microsoft Simple Encrypted Arithmetic Library (MS-SEAL). We significantly enhance the usability and applicability of FHE in deep learning contexts, with a focus on the constituent graphs, traversal, and optimisation. We find that FHE is not a panacea for all privacy preserving machine learning (PPML) problems, and that certain limitations still remain, such as model training. However we also find that in certain contexts FHE is well suited for computing completely private predictions with neural networks. The ability to privately compute sensitive problems more easily, while lowering the barriers to entry, can allow otherwise too-sensitive fields to begin advantaging themselves of performant third-party neural networks.  Lastly we show how encrypted deep learning can be applied to a sensitive real world problem in agri-food, i.e. strawberry yield forecasting, demonstrating competitive performance. We argue that the adoption of encrypted deep learning methods at scale could allow for a greater adoption of deep learning methodologies where privacy concerns exists, hence having a large positive potential impact within the agri-food sector and its journey to net zero.
    \end{abstract}
    
}

\maketitle

\IEEEdisplaynontitleabstractindextext

%
\IEEEpeerreviewmaketitle

\IEEEraisesectionheading{\section{Introduction}\label{sec:introduction}}

\begin{figure}[ht]
  \centering
  \includegraphics[width=\columnwidth]{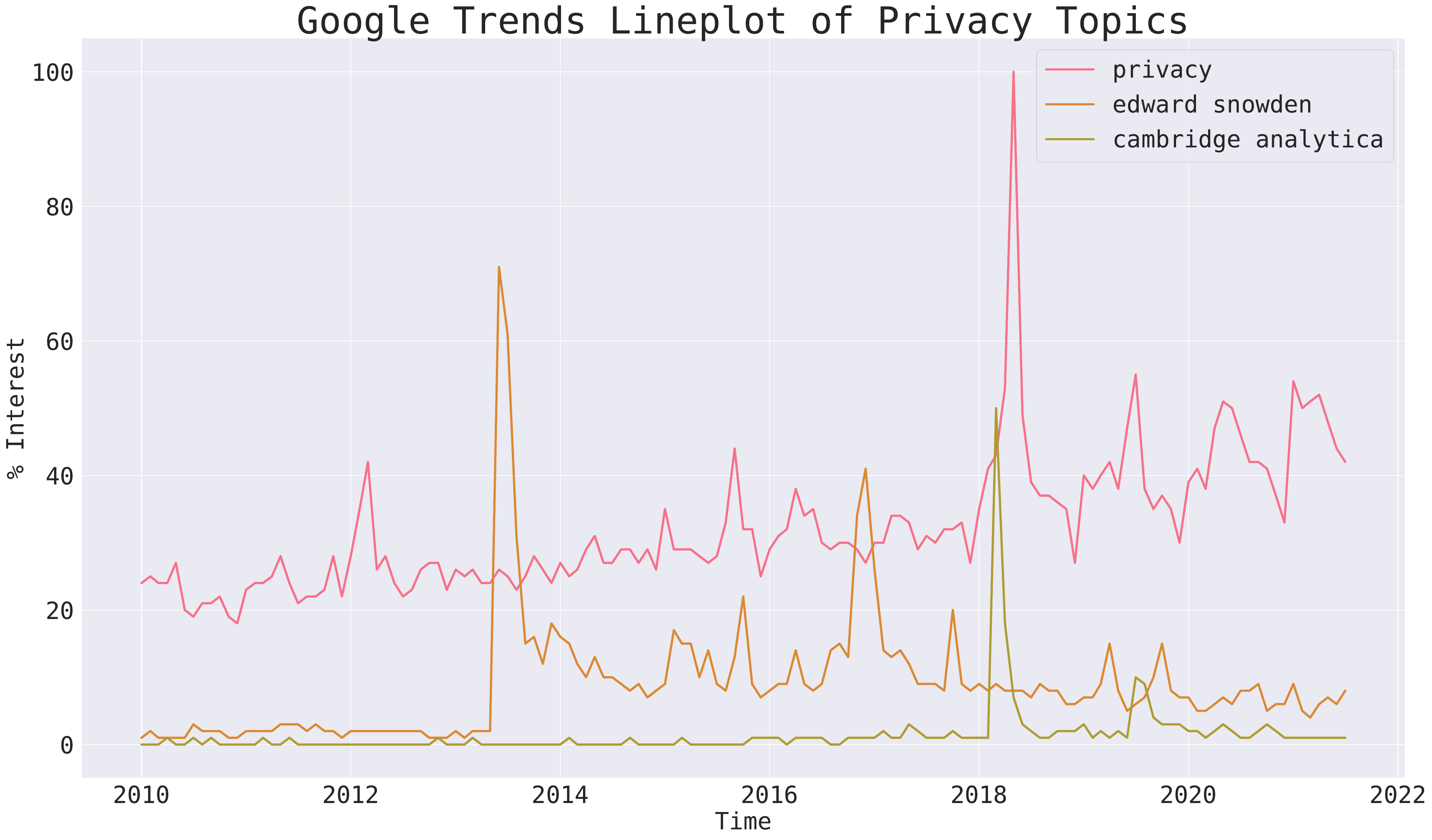}
  \caption{Trends of privacy (blue), Edward Snowden (red), and Cambridge Analytica topics (yellow) on Google trends since 2010 showing a slow but steady increase in the interest of privacy, and particular peaks around events such as the Cambridge Analytica scandal and smaller peaks roughly correlated to Julian Assange. \autocite{google-trends-privacy}}
  \label{fig:trends-privacy}
\end{figure}

%
%
%
%
\IEEEPARstart{F}{ully Homomorphic Encryption}
Privacy is slowly becoming of greater interest (Figure: \ref{fig:trends-privacy}) to the broader public, especially during and after particular scandals, such as Cambridge Analytica (corporate actors), Edward Snowden on the five eyes (state actors), \autocite{permanent_record} and more recently the Pegasus project on the cyberarms NSO group (both corporate and state). This increased concern for privacy has over time manifested itself in many forms; one of the most notable example being in legislation such as the General Data Protection Regulation (GDPR) \autocite{gdpr}.

A less thought-of field where privacy is of concern is the agri-food sector. Agriculturalists often are incredibly reluctant to share data, due to real, or perceived sensitivity. We believe that this data sharing reluctance originates from two factors; Data is not being collected due to the unawareness of the value-for-cost it can offer, and data is not shared due to concerns over loss of competitiveness if their techniques were leaked. This means it is incredibly difficult for new and possibly disruptive approaches to be used toward forecasting and thus later optimising some component in the agri-food chain. One such disruptive approach is the application of deep learning which has become state-of-the-art in almost all areas where sufficient data is present with which to train it. There are many reasons why such new approaches are necessary but the key area we gear our work towards is tackling food waste at production, by forecasting accurate yields.
Here in the UK we have dual problems of food insecurity and high food waste. It is estimated that the annual combined surplus and food-waste in primary production sis 3.6 million tonnes (Mt) or 6-7\% of total harvest. A further 9.5Mt is wasted post production / farm. 7.7Mt is wasted in house holds and 1.8Mt is wasted in manufacturing and retail. While the total food purchased for consumption in the UK is 43Mt. \autocite{foodSecurity21} More specifically in the soft-and-stone fruit industry a large consortium of growers in 2018 over estimated by 17.7\% for half of the growing season, while the remainder of the season they under-estimated by 10\%. Underestimation leads to surpluses which create extra cost in fruit disposal along with de-valuing expected produce. Overestimation leads to fix-purchasing which entails importing fruit to cover the shortfall in the expected produce. This costs the consortium 8 Million GBP a year in losses, while the rest of the industry is estimated to have incurred 18 Million GBP losses a year at the time. The effect of climate change has been exasperating the difficulties in yield forecasting due to the more erratic environmental conditions. Considering that freely available agri-food data are hard to find, given they are highly sensitive, progress in adopting AI technologies are hindered. 

As far as using machine learning is concerned, It is extremely difficult to build and deploy neural network models to forecast agricultural yields due to the aforementioned privacy/ sensitivity concerns that mean data for training and using these neural networks is scarce. However the impact of using machine learning technologies in agri-food supply chains has been shown to be substantial \autocite{kollias2022ai, onoufriou2019nemesyst, thota2021contrastive}. A solution that involved distributed learning was recently proposed with an application on soy bean yield forecasting \autocite{durrant2022role}, which assumes that distributed training is possible. Towards providing an alternative solution to this, we propose to use fully homomorphic encryption (FHE) and demonstrate how it works and performs in a bespoke strawberry dataset (Katerina and Zara varieties) that was collected in our strawberry research facility in Riseholme Campus at the University of Lincoln, UK.

FHE affords us the ability to compute cyphertexts without the ability to detect or discern its contents, acting as a truly blind data processor in Encrypted Deep Learning as a Service (EDLaaS) applications. \autocite{onoufriou2021fully} In particular EDLaaS is especially useful in highly sensitive/ highly regulated industries such as medicine/ patient data (especially due to GDPR), trade secrets, and military applications. FHE is not a panacea. Special care must be taken to ensure/ maximise the security of cyphertexts and the biggest problem with this is it is not immediately apparent if this is not ensured often requiring a deep understanding of the underlying cryptography such that the parameterisation can be understood, analysed, and balanced against. However a standard metric used thought as a commonality is the number of bits used for the private keys. It is commonly considered that a private key with 128 bits is considered secure. \autocite{seal, Dathathri_eva_2020} We maintain this minimum level of security thought all our experimentation and implementations.
\subsection{Contributions}
Our contributions towards FHE-deep-learning given the current state of the field and related works (Section: \ref{sec:related}) are:

\begin{itemize}
    \item To provide and showcase open-source encrypted deep learning with a reproducible step-by-step example on an open dataset, in this case Fashion-MNIST. Provided through a dockerised Jupyter-lab container, such that others can readily and easily explore FHE with deep learning and verify our results.
    \item To outline a new block-level automatic cyphertext parameterisation algorithm, which we call autoFHE. We also seek to showcase autoFHE in both regression and classification networks, which still appears to be a misunderstood and ongoing problem. \autocite{falcetta2022privacy}
    \item Show a new application for encrypted deep learning to a confidential real-world dataset.
    \item Detail how neuronal firing in multi-directed graphs can be achieved in our different approach.
    \item Show and detail precisely the computational graph of how a CNN can be constructed using FHE in particular how handling of the sum-of-products can occur. This along with our easily reproduced example, should help clarify many otherwise omitted details from previous works.
    \item Show recent advancements in FHE compatibility like ReLU approximations in greater detail along with problems/ considerations as part of a whole computational graph. We also backpropogate the dynamically approximate range of ReLU.
\end{itemize}

\section{Literature and Related Works}
 \subsection{FHE Background}
 FHE is a structure-preserving encryption transformation \autocite{gilad2016cryptonets}, proposed by Craig Gentry in 2009 \autocite{gentry2009fully}, allowing computation on cyphertexts ($\varepsilon(x)$) directly (addition and multiplication) without the need for decryption. This is what could be considered the first generation of FHE as implemented by Gentry in 2011 \autocite{gentrysFirstFhe} and the Smart-Vercauteren implementation \autocite{smart-vercauteren}. Gentry's implementation for any given bootstrapping operating took anywhere from 30 seconds, for the smallest most "toy" example, to 30 minutes for the largest most secure example, with the former having a public-key of 70 Megabytes, and the latter a public-key of 2.4 Gigabytes in size. \autocite{gentrysFirstFhe}
Clearly this would be far too lengthy to be practically viable, however there have been several generations of FHE since building on these initial works and improving computational and spacial complexity; Second generation: BV \autocite{bv}, BGV \autocite{bgv}, LTV \autocite{ltv}, BFV \autocite{bfv}, BLLN \autocite{blln}; Third generation: GSW \autocite{gsw}; Fourth generation: CKKS \autocite{ckks}.
Here we focus on the Cheon, Kim, Kim and Song (CKKS) scheme, for a plethora of reasons:
\begin{itemize}
    \item Operates with fixed point precision unlike all other schemes, which are necessary for computation of neural networks with activations and inputs usually falling in the range $0, \pm1$ \autocite{ckks2018}
    \item Has multiple available implementations (PALISADE \autocite{al2022openfhe}, HEAAN \autocite{cheon2017homomorphic}, Microsoft Simple Encrypted Arithmetic Library (MS-SEAL) \autocite{seal}, HElib \autocite{halevi2020design}, etc). Only PALISADE \autocite{al2022openfhe} and Lattigo \autocite{mouchet2020multiparty} are known to implement CKKS with bootstrapping, although many others have these features road-mapped.
\end{itemize}

Our implementation uses MS-Seal, an incredibly popular FHE library. Many of our techniques proposed here stretch to almost all other implementations since they follow the same basic rules, albeit with slightly different implications on things like parameters. We are also in the process of translating our work to other implementations (Darklantern \autocite{darklantern} using Lattigo \autocite{mouchet2020multiparty}) to advantage our work of things such as bootstrapping.
Bootstrapping can afford us longer-effective viable-computational-depth. Thus our work here focuses on using FHE without bootstrapping, or more precisely levelled-fully-homomorphic-encryption (LFHE), meaning we calculate specific sized although generalized (implementation) neural network circuits. Despite CKKS being the best candidate for forms of encrypted deep learning, it has certain shortcomings.
  
\begin{figure}[t!]
  \centering
  \includegraphics[width=\columnwidth]{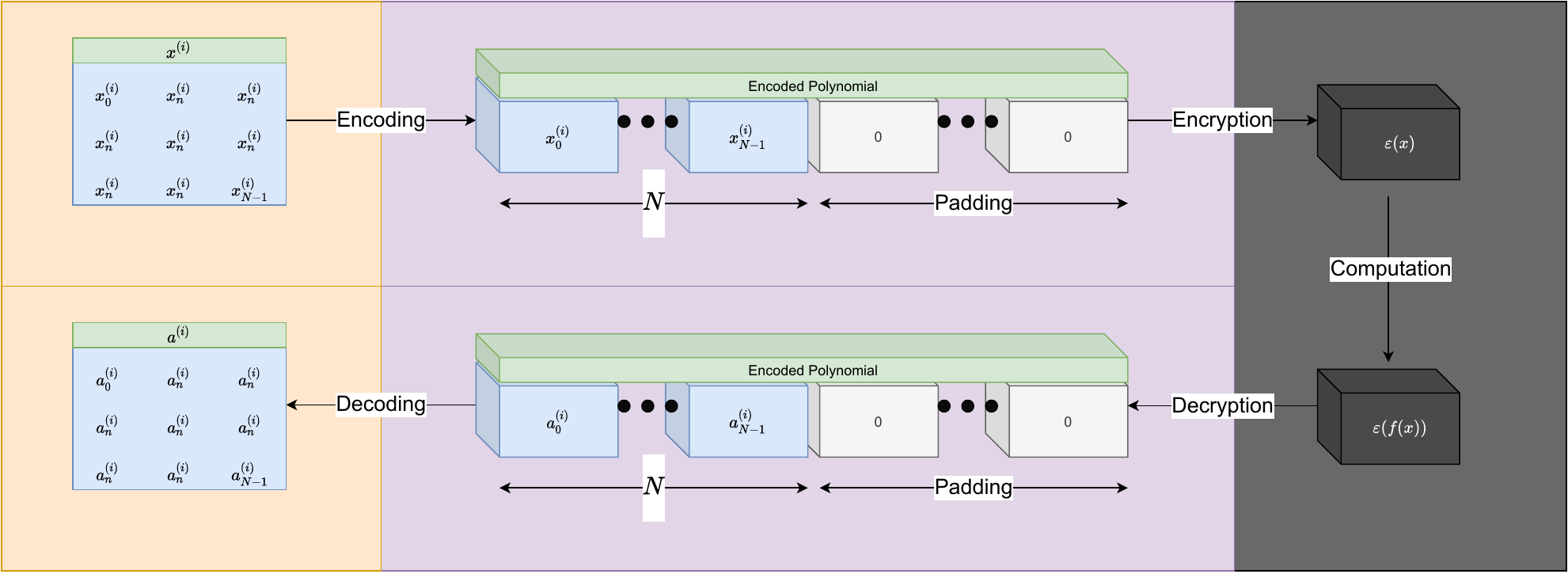}
  \caption{Fully Homomorphic Encryption (FHE) overview of distinct stages and properties. \autocite{reseal}}
  \label{fig:fhe}
\end{figure}

  CKKS cyphertexts are the most atomic form of the data. This is a consequence from the optimization used in many FHE schemes where a sequence of values (the "message" or plaintext data) are encoded into a single polynomial, and then this polynomial is what is then encrypted (Figure: \ref{fig:fhe}). This means there is less overhead since we are encrypting multiple values together, but it means we cannot operate on this value alone, we must always be homomorphic, i.e maintain the same structure and operate on all values. Thus if we encrypt a polynomial of length 10, that shall be the smallest form of the data until it is either bootstrapped or re-encrypted. We are hence only able to operate on the 10 elements as a single whole, E.G we cannot operate on the 3rd element in the array alone to produce a single number answer.
  
  CKKS cyphertexts computational depth (pre-bootstrapping) is directly related to the length of the polynomial slots. Which means we must choose our parameters carefully to ensure we do not have unnecessarily large cyphertexts, and thus slow operations.
  
  CKKS as with many schemes requires that two cyphertexts operating with each other, must share the same parameters and be from the same private key. This means when for instance we have multiple inputs into a neural network, all directly interacting cyphertexts must be of the same key. This complicates some automatic parameterisation logic which we will discuss later.

\subsection{Related Works}\label{sec:related}
\subsubsection{Encrypted Deep Learning}
There have been many other works that use FHE (bootstrappable) or Levelled-FHE to compute some form of neural network. A few notable examples for FHE and Convolutional Neural Networks (CNNs) are by Lee, \autocite{lee2021privacy}, Meftah, \autocite{meftah2021doren}, Juvekar \autocite{juvekar2018gazelle}, and Marcano, \autocite{marcano2019fully}. Lee uses a modified version of the Microsoft Simple Encrypted Arithmetic Library (MS-SEAL) to add bootstrapping as MS-SEAL does not currently support it. Lee shows FHE and DL used on the CIFAR-10 \autocite{cifar10} dataset to mimic the ResNet-20 model achieving a classification accuracy of 90.67\%. Juvekar uses the PALISADE library implementation of the BFV scheme with their own (LFHE) packed additive (PAHE) neural network framework to compute both MNIST and CIFAR-10.  Meftah uses Homomorphic Encryption Library (HELib) \autocite{halevi2020design} similarly to Lee is particularly focused on improving the practicality of (L)FHE as a means to compute deep learning circuit. Meftah seeks to do this towards computing ImageNet \autocite{deng2009imagenet} with the second generation BGV scheme \autocite{bgv} (on integers) as opposed to Lee using the fourth generation CKKS scheme \autocite{ckks} (on floating points). Lastly Marcano similarly to the previous is also concerned with the computational, and spatial complexity of using FHE as a means to compute convolutional circuits. Marcano appears to use a custom FHE implementation on fixed point number format, taking 36 hours to train on the MNIST dataset. It is unclear in all of these papers however, how exactly the gradient descent or backward pass of the neural networks are implemented, which is necessary for neural network training. They also lack detail in key stages of the forward pass such as how they dealt with calculating the sum-of-products of the CNN since a homomorphic cyphertext cannot be folded on itself to form a single number sum, or if they used point-wise encryption to be able to sum between cyphertexts how they dealt with the sheer size of this plethora of cyphertexts. Lastly the above papers do describe in some detail how some of their parameters are decided in particular with regards to security, but they do not cover much on the computational depth or precision effects these parameters have on the cyphertext such as the modulus-switching chain.
\subsubsection{FHE Graph Parameterisation}
Here FHE graph parameterisation means deriving the FHE parameters from a graph, such as the computational depth and thus the parameters like the modulus size. There have been a few works that define FHE graph parameterisation, the most notable and similar of which is Microsoft Encrypted Vector Arithmetic (MS-EVA) \autocite{Dathathri_eva_2020, falcetta2022privacy}. MS-EVA uses Directed Acyclic Graphs (DAGs) to represent simple operations applied to some input constant. Since MS-EVA also uses MS-SEAL this means it also uses RNS-CKKS the purportedly most efficient CKKS implementation. \autocite{Dathathri_eva_2020} MS-EVA has been applied to encrypted deep learning inference, specifically LeNet-5 towards MNIST. Dathathri particularly emphasises the non-trivial nature and how parameterisation can be a large barrier to the adoption of FHE. However there are no examples currently available to help lower this barrier. Subsequently their nodes representing single atomic operations means there is overhead when compared to block operations which could be an area of improvement.

\subsection{Threat Model}

Just like similar works in FHE we assume a semi-honest/ honest-but-curious threat model. \autocite{Dathathri_eva_2020} Where parties follow the specified protocol but attempt to garner as much possible information from their received messages as possible. Or indeed one party has malicious intrusion which can read the data shared, but not necessarily write/ change the protocol.

\section{Fully Homomorphic Encryption and Deep Learning}
As a necessary pre-requisite there is some prior understanding about FHE that is necessary but not broadly well known in particular when applied to deep neural network graphs that are often seen in the field of deep learning. We would like to highlight those here to make it clear in other sections how we overcome these limitations and highlight the advancements we make here. We would also like to note that Fully Homomorphic Encryption as a concept is distinct from any specific implementation scheme as we have previously eluded to. In our case the scheme we use is the Cheon Kim Kim Song (CKKS) scheme as previously stated and described, however following is some further information that applies to this scheme:

\begin{itemize}
    \item Two cyphertexts that operate together must be identical containers; Same scheme, the same size, the number of primes into their swapping chain, and they must originate from the same private key.
    \item Additions double the noise of a cyphertext whereas a multiplication exponentially increases the noise, which means to reduce the noise we must consume an element in our swapping chain to reduce the noise again. Since multiplication is much noisier than addition we tend to only swap after multiplication.
    \item Abelian compatible operations are the only operations that can occur on an FHE cyphertext. This means addition and multiplication. There are methods to model division and subtraction but these operations are impossible under FHE. Thus the need to create new methods and algorithms.
    \item Cyphertexts size and number of primes in the swapping chain are related. The bigger the cyphertext the more primes it contains for swapping. However the bigger the cyphertext the longer the computation takes. Thus we want the smallest possible cyphertext that has enough primes to complete the set amount of computations.
    \item Cyphertexts of a larger size also contain more slots, these slots are what are use to store our message or input/ plaintext data. Thus we must also consider that to store a certain number of features we must have a certain sized cyphertext. The CKKS scheme has half the number of slots compared to other schemes for the same size since it models pre and post point fixed precision.
    \item Once the swapping chain has been consumed a very expensive operation called bootstrapping is necessary to refresh the cyphertext and regenerate the swapping chain to continue to do noise-expensive operations.
    \item If the cyphertext is too noisy at the point of decryption it will lose precision or if even more noise is present the decrypted message/ data will become garbled and incorrect.
\end{itemize}

All of these points must be considered in the implementation of FHE compatible neural networks, and this is the primary reason why most existing work in the deep learning field is unfit for use under FHE including existing deep learning libraries.

We would also like to highlight as a consequence that there is little work in the domain of FHE deep learning with which to compare to and draw techniques from.

\section{Materials and Methods}
\begin{algorithm}[h]
\caption{Neuronal-Firing, our exhaustive neuron stimulating, depth-first, blocking, node-centric, graph/ neuron stimulation function.}\label{alg:firing}
\begin{algorithmic}
\Require{$g$: Neural network multi-directed computational graph}
\Require{$n$: Vector of neurons/ computational nodes for sequential stimulation}
\Require{$s$: Vector of signals to be induced in the corresponding neuron}
\Require{$r$: Vector of receptors to call on respective node}
\Ensure{$g'$: Stimulated neural network/ modified computational graph}
 \For{$i \gets 0\ \text{to}\ $\Call{length}{$n$}}
  \State \Call{signal\_carrier}{$g, n[i], r[i], s[i]$}
 \EndFor
\end{algorithmic}
\end{algorithm}

\begin{algorithm}[h]
\caption{Neuronal-firing-signal-carrier; Propagate a single signal thought all possible nodes in the neural network graph recursively based on its position.}\label{alg:carry_signal}
\begin{algorithmic}
 \Function{signal\_carrier}{$g,\ n,\ r,\ \textit{bootstrap}$}
  \State $s \gets$\Call{get\_inbound\_signal}{$g, n, r, \textit{bootstrap}$}
  \If{$s = \textit{None}$}
    \State \Return{$\textit{None}$}
  \EndIf
  \State $s \gets$\Call{apply\_signal}{$g, n, r, s$}
  \If{$s = \textit{None}$}
    \State \Return{$\textit{None}$}
  \EndIf
  \State \Call{set\_outbound\_signals}{$g, n, r, s$}
 \EndFunction
 \ForAll{$\textit{successors}\ \text{in}\ g\text{.node}(n)\text{.successors}()$}
 \State \Call{signal\_carrier}{$g,n,r,\textit{None}$}
 \EndFor
\end{algorithmic}
\end{algorithm}

\begin{algorithm}[h]
\caption{Get accumulated inbound signal from edges.}\label{alg:get_signal}
\begin{algorithmic}
 \Function{get\_inbound\_signal}{$g$, $n$, $r$, $\textit{bootstrap}$}
  \If{$\textit{bootstrap} \neq \textit{None}$}
    \State \Return{$\textit{bootstrap}$}
  \EndIf
  \State $s\ \gets\ [\ ]$
  \ForAll{$\text{edges}\ \text{in}\ g\text{.in\_edges(}n\text{)}$}
    \State $s\text{.append}(\text{edge.signal}(r))$
  \EndFor
  \If{$\text{length}(s) = 1$}
    \State \Return{$s[0]$}
  \EndIf
  \Return{$s$}
 \EndFunction
\end{algorithmic}
\end{algorithm}

\begin{algorithm}[h]
\caption{Activate current node using the accumulated signal and get outbound signal.}\label{alg:apply_signal}
\begin{algorithmic}
 \Function{apply\_signal}{$g$, $n$, $r$, $s$}
  \If{$s = \textit{None}$}
    \State \Return{$\textit{None}$}
  \EndIf
  \State $s \gets g \text{.nodes}(n)\text{.receptor}(r,\ s)$
  \State \Return{$s$}
 \EndFunction
\end{algorithmic}
\end{algorithm}

\begin{algorithm}[h]
\caption{Set outbound edges with activation signal.}\label{alg:set_signal}
\begin{algorithmic}
 \Function{set\_outbound\_signals}{$g$, $n$, $r$, $s$}
  \If{$s = \textit{None}$}
    \State \Return{$\textit{None}$}
  \EndIf
  \ForAll{$\text{edges}\ \text{in}\ g\text{.out\_edges(}n\text{)}$}
    \If{$\text{isinstance}(s,\ \textit{generator})$}
      \State $\text{edge.signal}(r) \gets next(s)$
    \Else
      \State $\text{edge.signal}(r) \gets s$
    \EndIf
  \EndFor
 \EndFunction
\end{algorithmic}
\end{algorithm}

To enable this research it was necessary to create our own python-based FHE compatible deep learning library because there was still a significant lack of compatibility between existing deep learning libraries and existing FHE libraries. While it may be possible to create some form of interface or bridge this left much to be desired in terms of usability and flexibility to explore different research avenues like various FHE backends. As a consequence we created a NumPy API focused library, where the inputs to the neural networks need only conform to the basic NumPy custom containers specification, allowing the objects passed in to handle their own nature. This means any NumPy conforming object can be used in our networks, this includes NumPy itself (for pure plaintexts) or in this case arbitrary FHE objects.
Our research here focuses on CPU computations as compatibility with existing CUDA implementations is currently infeasible due to compatibility which means conducting FHE over GPUs would be extremely difficult at this time.
Encrypted deep learning accelerated by GPUs is an area we seek to explore in the future, for the rest of this paper however all operations are conducted on CPUs. Our entire source code for our library Python-FHEz is available online along with the respective documentation. \cite{reseal} We use the MS-SEAL C++ library bound to python using community pybind11 bindings to provide us with the necessary FHE primitives which we then wrap in the NumPy custom container specification for the aforementioned reasons. \cite{huelse}\\

Furthermore in this section we outline our specific implementation, techniques, equations, and methods used to exemplify EDLaaS in practice using both an open dataset, and a preview of more real-world/ complicated but proprietary data scenario. We do this to enable some comparisons to be drawn and to introduce an new way of solving problems encountered in the agri-food industry:
\begin{itemize}
    \item We chose to use Fashion-MNIST, consisting of a training set of 60,000 examples and a test set of 10,000 examples as our classification example as it is a drop in replacement for the MNIST dataset while being more complex, but still familiar to most.
    \item We also chose to use an agri-food but proprietary dataset to exemplify a different kind of regression network and how FHE might play a role in this sensitive industry where data sharing/ availability is scarce due to a barrier in concerns over competition, of which FHE might help reduce \cite{pearson2019distributed}. Agri-food is also a key industry which has had a troubled few years due to climate change bringing hotter/ record-breaking summers, while also being effected by both coronavirus and Brexit shortages in staffing and thus supplying. In addition, it has been established that data sharing is a hindering factor that prevents machine learning technologies from being adopted at scale \cite{durrant2021might} but some work has already been done around using federated learning to alleviate some of these issues \cite{durrant2022role}.
\end{itemize}

For our neural networks we used a node-centric, multi-directed graph approach where:
\begin{itemize}
  \item Each node represents some computation object usually a neuron.
  \item Each edge represents the movement of data between neurons/ computation objects.
  \item Each node can accept many inputs that are stacked on top of each other in the same order as the edges, unless there is a single input edge where it is instead mapped to the input of the neuron.
  \item Each edge can only connect two nodes directed from the first to the second node, parallel edges are possible and are treated as completely separate edges with no special handling.
  \item A node can only be activated/ computed once all predecessor edges carry some data.
  \item All nodes can have several receptors, that is to say different functions that can be pointed to by the edges, in particular forward and backward receptors for calculating the forward neural network pass, and gradients using the chain rule in the backward pass.
  \item Nodes return either an iterable to be equally broadcast to all successor edges or a generator to generate independent results for each successor edge.
  \item The weight of each edge corresponds to the computational depth of the directed-to node. These weights are not used to optimise the path since the majority of nodes must be activated to achieve some desired output, but instead these weights are used to find the longest path between key-rotations to determine the minimum required encryption parameters to traverse from one rotation to the next.
  \item Self-loop edges are not treated differently, instead relying on the configuration of the node itself to consider termination of the loop.
  \item A single activation pass of the graph may have multiple input and multiple output nodes/ neurons, like in the two blue regions in the sphira graph (Figure: \ref{fig:fashion_mnist_cg}).
\end{itemize}

We do this from the node perspective as we find this to be more conceptually clear and follows our own mental abstractions of how neural networks operate. This makes it easier for us to conceptualise, implement, and communicate our neural networks, in particular visually.

To activate our neural network graph we used our own neuronal-firing algorithm (Algorithm: \ref{alg:firing}), since we could not find better existing algorithms that would be suitable for firing of encrypted neuron graphs, while offering us the flexibility to adapt to changing our research.

\subsection{FHE parameterisation}

\begin{algorithm}[t]
\caption{Automatic FHE-parameterisation by source and cost discovery, over multi-directed graphs.}\label{alg:autohe}
\begin{algorithmic}

\Function{autoHE}{$g$, $n$, $\textit{concern}$}
    \For{$i\ \text{in}\ n$}\Comment{Label graph sources and costs}
        \State $\text{autoHE\_discover}(g, i, i, \textit{concern}, 0)$
    \EndFor
    \State $r\ \gets\ \text{tuple}(\text{dictionary()}, \text{list()})$\Comment{Group representation}
    \For{$i\ \text{in}\ n$}\Comment{Assign + merge groups from labels}
        \If{$r[0]\text{.get}(i)\ \text{is}\ \textit{None}$}
            \State $r[0][i]\ \gets\ \text{len}(r[1])$
            \State $r[1]\text{.append}(0)$
        \EndIf
        \For{$j\ \text{in}\ g\text{.nodes}()$}
            \State $\textit{src}\ \gets\ j[1][\text{"sources"}]$
            \If{$i\ \text{in}\ \textit{src}$}
                \For{$k\ \text{in}\ \textit{src}$}
                    \State $r[0][k]\ \gets\ r[0][i]$
                    \If{$\textit{src}[k]\ >\ r[1][r[0][i]]$}
                        \State $r[1][r[0][i]] = \textit{src}[k]$
                    \EndIf
                \EndFor
            \EndIf
        \EndFor
    \EndFor
  \Return{$r$}

\EndFunction
\end{algorithmic}
\end{algorithm}

\begin{algorithm}[t]
\caption{Recursive FHE-parameterisation source, and cost discovery over multi-directed graphs.}\label{alg:autohe_discover}
\begin{algorithmic}

\Function{autoHE\_discover}{$g$, $n$, $s$, $\textit{concern}$, $c$}
  \State $d\ \gets\ g\text{.nodes}()\text{.get}(n)$
  \If{$d\text{.get("sources")} = \textit{None}$}
      \State $d[\text{"sources"}]\ \gets\ \text{dict}()$
  \EndIf
  \If{$s\ \neq\ n$}
      \If{$d[\text{"sources"}]\text{.get}(s)\ =\ \textit{None}$}
          \State $d[\text{"sources"}][s]\ \gets\ c$
      \ElsIf{$d[\text{"sources"}]\text{.get}(s)\ <\ \textit{c}$}
          \State $d[\text{"sources"}][s]\ \gets\ c$
      \EndIf
  \EndIf
  \If{$\text{isinstance}(d[\text{"node"}], \textit{concern})$}
      \State $\text{autoHE\_discover}(g, n, n, \textit{concern}, 0)$
  \Else
      \For{$i\ \text{in}\ g\text{.successors}(n)$}
        \State $\textit{nxt}\ \gets\ c + g\text{.nodes}()[i][\text{"node"}]\text{.cost}()$
        \State $\text{autoHE\_discover}(g, i, s, \textit{concern}, \textit{nxt})$
      \EndFor
  \EndIf
 \EndFunction
\end{algorithmic}
\end{algorithm}

Our automatic FHE parameterisation approach is similar to that of MS-EVA \autocite{Dathathri_eva_2020} where we use (in our case our existing neural network) graphs to represent the computation the cyphertexts will experience. This allows us to automatically generate the smallest secure cyphertext possible that meets the requirements of the proceeding computational circuit. How we differ however is that since we are using neural network neurons instead of atomic (addition, multiplication, etc) operations, there are fewer nodes and edges, and thus less overhead necessary of both the graph, and any intermediate storage along edges. This is because we can block-optimise at a higher level that would be possible if purely considering individual atomic operations. Also our neural network graphs are Multi Directed Graphs (MDGs) as opposed to Directed Acrycling Graphs (DAGs) which means we can model more complex operations involving more than two inputs. This affords us the ability to model the complex relationships in neural networks much like standard deep learning libraries.

In our abstraction automatic FHE parameterisation becomes a variation of the travelling-salesman problem, but instead of finding the shortest path we need to find the longest possible path or more specifically the highest computational depth experienced by the cyphertext, between sources and sinks. However even in our abstraction we must still conform to the constraints of CKKS, like that interacting cyphertexts must match, in cyphertext scales, and must be originating of the same private key which means other adjoining paths must be considered where they intersect. A key distinction compared to MS-EVA's approach is that our graphs are interpreted instead of being compiled down to some intermediate representation. Our cyphertext objects are also not raw, and are instead part of a larger NumPy-API compatible objects that interpret invocations. These meta objects are also responsible for the decision making of both relinearisation and re-scaling, taking that complexity away from the implementation of encrypted deep learning. An example of this rescaling interpretation is when two cyphertexts are multiplied, the meta-object is responsible for ensuring both cyphertexts match, I.E swapping down the modulus chain to equal scales depending on which of the two cyphertexts is higher up the modulus switching chain. Similarly an example of relinearisation is when two of our meta-objects are multiplied the computing member (usually the first meta-object in sequence) automatically relinearises the new meta-object, before passing the new meta-result back. This means we offload re-scaling and relinearisation, and it is not necessary to plan for these two operations, instead we need only calculate the longest paths, and the "groups" of cyphertexts. Here groups of cyphertexts means cyphertexts that interact, and must then share encryption parameters.

In short the minimum necessary information we need to derive from the graph using our algorithms (Algorithms: \ref{alg:autohe}, \ref{alg:autohe_discover}) is:
\begin{itemize}
    \item Which cyphertexts interact at which nodes
    \item Thus which nodes belong to which group
    \item What is the maximum computational depth of each group necessary to go from one (type-of-concern) source to another (type-of-concern) sink/ rotation
\end{itemize}

\begin{figure}[t!]
  \centering
  \includegraphics[width=\columnwidth]{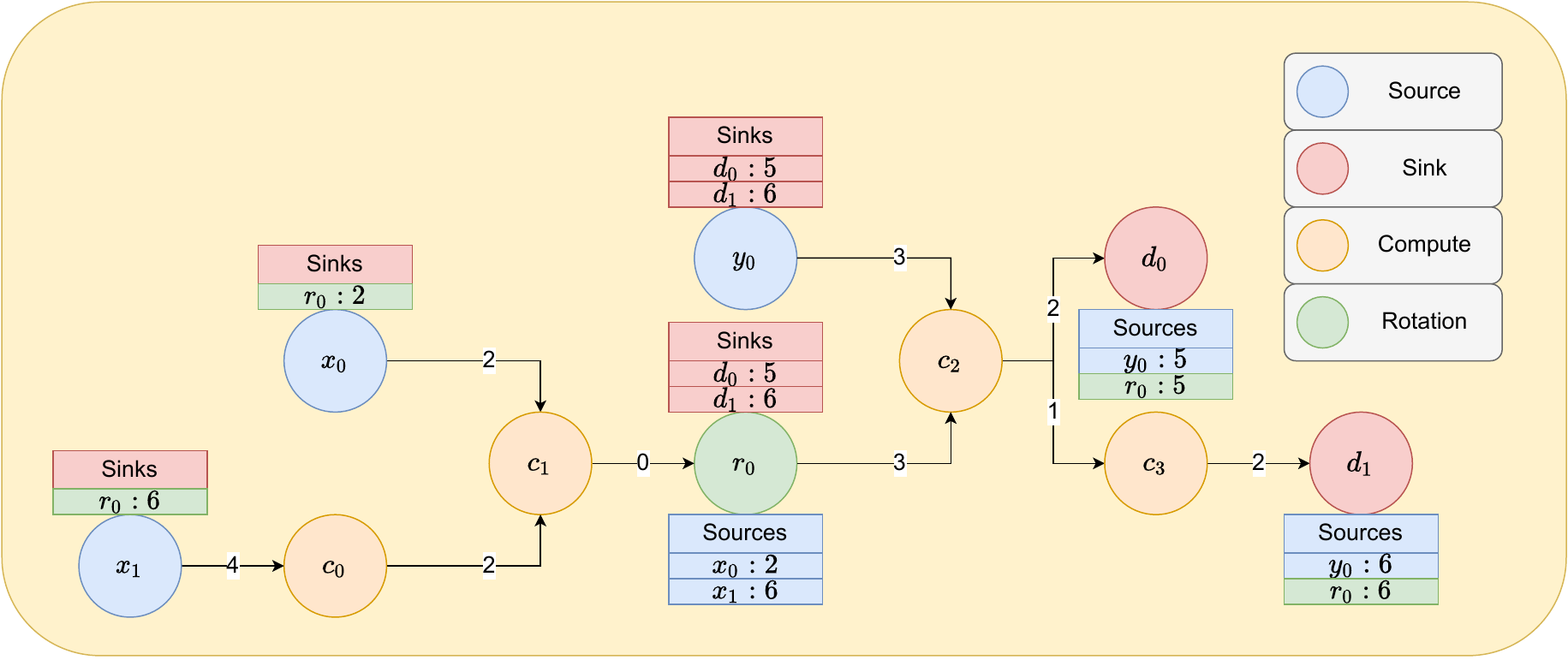}
  \caption{Example automatic FHE parameterisation problem, over a multi-directed graph. Sources are where data becomes a cyphertext. Sinks are where cyphertexts become plaintexts. Computation nodes are generic nodes that represent some operation that can be applied to both cyphertexts and plaintexts. Explicit rotation nodes are where a cyphertexts keys are rotated, either to refresh them, or to change the form of the cyphertext, potentially into multiple smaller cyphertexts. Please note this does not necessarily follow the colour coding of our other automatically-generated graphs. \autocite{reseal}}
  \label{fig:fhe_parm_problem}
\end{figure}

Each of our nodes must be labelled with its computational depth, so that the highest-cost traversal can take place. This may need to occur multiple times in a single graph, depending on the number of sources and sinks in said graph. The more sources to sinks, the more paths the more intersections and co-dependency. Take for instance $x_0$, and $x_1$ in the dummy network depicted in Figure: \ref{fig:fhe_parm_problem}. The cyphertexts $x_0$ and $x_1$ passed in must be able to reach the end of both paths leading to $r_0$ the very next sink/rotation. To do this they must be inter-operable with each-other at the point at which they meet. This means they must have matching scales, encryption parameters, and must originate from the same private key. However consider that $x_1$ experiences computations $c_0$ and $c_1$ whereas $x_0$ only experiences $c_1$. Each computation changes the scale, and thus necessarily their remaining primes in the modulus switching chain which would make them inoperable if not for our specialised logic in the meta-object to match them automatically. For instance, spatial and temporal data in the case of multi-modal datasets (of which Fashion-MNIST is not) would have multiple inputs that require matching.
Since decisions on relinearisation and rescaling are left to the meta-object the only information we need to ordain from the graph is the computational depth, and co-dependency of parameters. This can then be used to associate parameters together and select the minimum viable polynomial modulus degree.

In our node centric view of the graphs we say an edge from node A to B has the cost associated with B. This algorithm should be able to handle multiple cyphertext ingress nodes ($x$, $y$, etc), multiple cyphertext egress nodes ($y_hat$, and any others), and key-rotation stages in-between that will also need to be parameterised along the way. Our proposed algorithm can be seen in Algorithm: \ref{alg:autohe}. The output of this algorithm is a tuple representation of the graphs parameterisation-groups. We will know which nodes need to share parameters, and what the highest cost of that parameter-group is. If we combine this graph parameter representation and some basic logic, we can tune/ parameterise automatically. This will of course vary for each implementation of FHE, from CKKS to BFV for example, requiring different parameters. The difference in parameterisation is why we separate out this final step, so that custom functions can be injected.

The FHE parameters we deal with here primarily geared toward the MS-SEAL CKKS backend are:
\begin{itemize}
  \item Scale; Computational scale/ fixed point precision
  \item Polynomial modulus degree; polynomial degree with which to encode the plaintext message, this dictates the number of available slots, and the available number total bits which the coefficient modulus chain can contain.
  \item Coefficient modulus chain; a list of byte sizes with which to switch down the modulus chain, this dictates the computational depth available before bootstrapping or key-rotation is necessary.
\end{itemize}
However the information we derive from the graph is generic and can be broadly adapted to generate parameters for other schemes also.

We use the default 128-bit security level of MS-SEAL just as MS-EVA \autocite{Dathathri_eva_2020}, being the most similar existing framework. This is security level is broadly considered reasonably secure \autocite{lee2021privacy, meftah2021doren, Dathathri_eva_2020}, and matches our threat model of honest-but-curious.

Lastly now that we have calculated the groups, the cost of the groups, and the associated nodes that belong to which groups, we can use a rough heuristic (Figure: \ref{alg:parameterise}) to estimate the necessary FHE parameters to accompany these groups. This heuristic can be tuned, and overridden for other FHE schemes to more tightly parameterise if necessary.

\begin{algorithm}[t]
\caption{Heuristically parameterise RNS-CKKS scheme using expected cost of computation.}\label{alg:parameterise}
\begin{algorithmic}
\Require{$c$: Integer maximal-cost of this cyphertext group.}
\Require{$s$: Integer scale-power, the scale of the cyphertext. Default: $s=40$. We advise not to go below $30$ due to noise accumulation and lack of prime availability.}
\Require{$p$: Float special-prime-multiplier, the multiplier that dictates the scale-stabilised special-primes in the coefficient-modulus chain. Default: $p=1.5$}
\Ensure{$\textit{parms}$: MS-SEAL RNS-CKKS parameter dictionary/ map.}
\Function{parameterise}{$c$, $s$, $p$}
    \State $\textit{parms}\ \gets\ \text{dict}()$
    \State $\textit{parms}[\text{"scheme"}]\ \gets\ 2$\Comment{2 is CKKS in MS-SEAL}
    \State $\textit{parms}[\text{"scale"}]\ \gets\ \text{pow}(2, s)$\Comment{scale power}
    \State $m\ \gets\ [s\ \text{for}\ i\ \text{in range}(c+2)]$
    \State $m[0]\ \gets\ \text{int}(m[0]*p)$\Comment{Mult first special prime}
    \State $m[-1]\ \gets\ \text{int}(m[-1]*p)$\Comment{Mult last special prime}
    \State $b\ \gets\ 27$
    \While{$b\ <\ \text{sum}(m)$}
      \State $b\ \gets\ b*2$
    \EndWhile
    \State $\textit{parms}[\text{"poly\_modulus\_degree"}]\ \gets\ \text{int}(1024 * (b/27))$

    \State \Return{$\textit{parms}$}
\EndFunction
\end{algorithmic}
\end{algorithm}

\subsection{Open Data Fashion-MNIST}

\begin{figure}[t!]
  \centering
  \includegraphics[width=0.3\columnwidth]{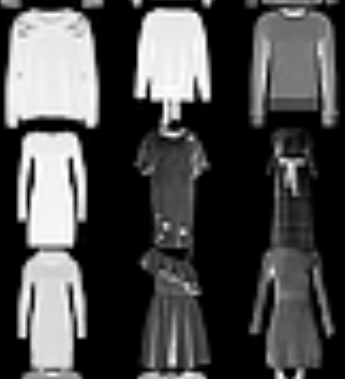}
  \caption{Fashion-MNIST sample showing examples of data such as: boots, bags, jumpers, and trousers. \autocite{DBLP:journals/corr/abs-1708-07747}}
  \label{fig:fashion_mnist}
\end{figure}

\begin{figure}[h]
  \centering
  \includegraphics[width=0.8\columnwidth]{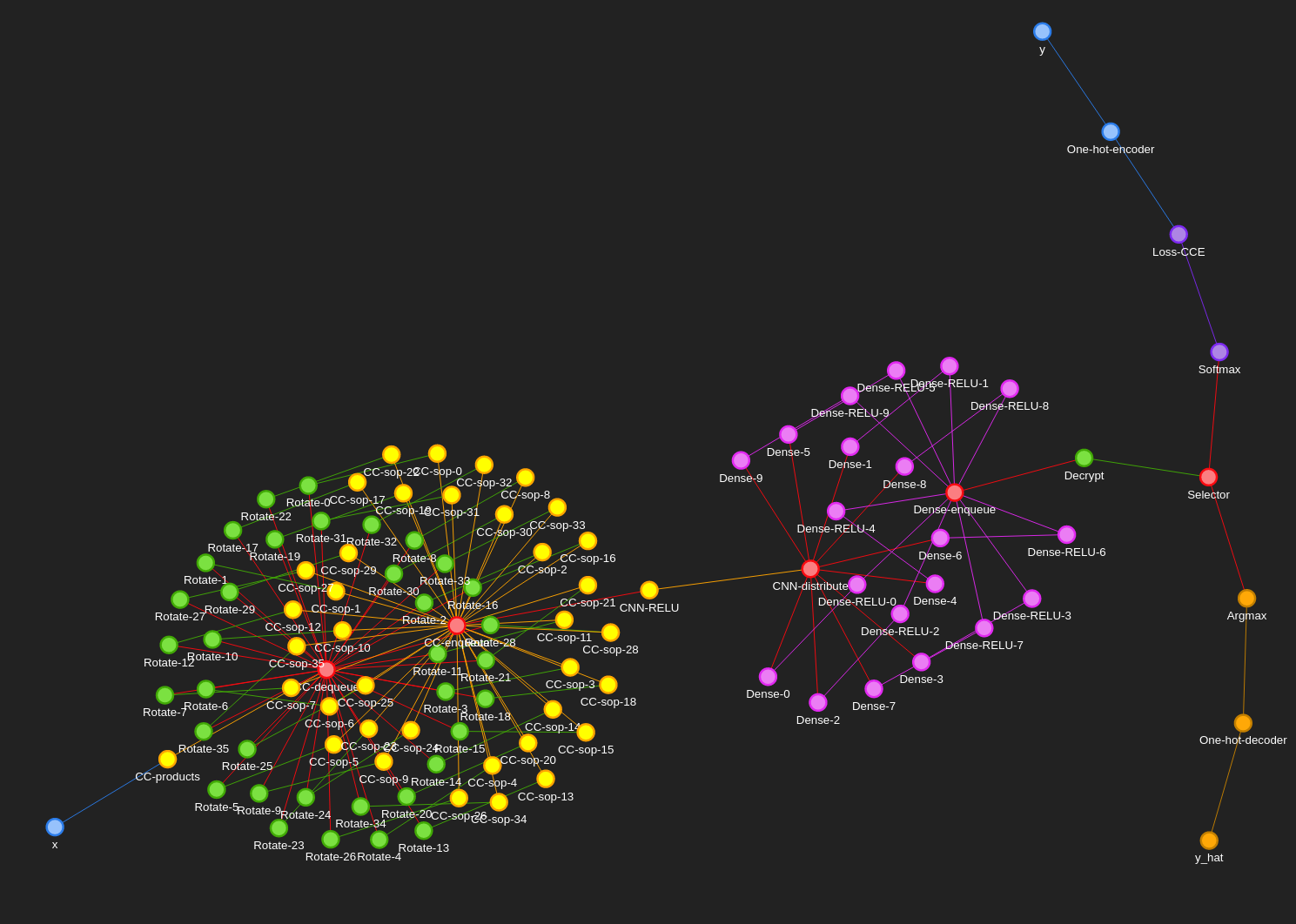}
  \caption{Fashion-MNIST computational graph we call "sphira", showing the colour coded graph and the respective nodes used to train/ compute Fashion-MNIST using our neuronal-firing algorithm. Blue represents the input and input transformation circuit that deals with passing the signals into the neural network in a way it is expecting them. Yellow represents the convolutional neural network components where one filter neuron passes multiple output cyphertexts to a plethora of summing nodes. Pink represents the fully connected dense layer for each class. Purple represents the loss calculation circuit necessary for backpropagation. Orange represents the output/ prediction circuit. Red represents the generic glue operations necessary to bind components together. Green represents the encryption specific nodes like decryption, rotation, encryption. An interactive version of this graph is available in our source code documentation so that clusters of nodes can be peeled apart for investigating individual nodes and connections. \autocite{reseal}}
  \label{fig:fashion_mnist_cg}
\end{figure}

In this section we describe our openly available Jupyter implementation \autocite{reseal} of an FHE-compatible CNN operating on the open dataset called Fashion-MNIST as can be seen in Figure: \ref{fig:fashion_mnist}. This dataset contains in total 70,000 images, 60,000 for training and 10,000 for testing. This dataset contains images of certain items of clothing, constituting 10 classes. Each image is a mere 28x28x1 pixels.
The full implementation can be found in the examples of our source code repository. \autocite{reseal}

We chose Fashion-MNIST as it is a drop in replacement for MNIST while also being a somewhat more difficult problem than standard MNIST. Coincidentally being that MNIST and thus Fashion-MNIST are both classification rather than regression they represent an even more difficult scenario for encrypted deep learning since they do not provide one continuous/ regressed output so the computational circuit becomes more complex/ deeper as far as necessary to process these classifications I.E the extra dense nodes for each class, and the whole addition of both softmax and categorical cross entropy (CCE) to replace the mean squared error (MSE) loss function in the case of would be regression networks. This also poses a problem as methods usually used towards classification like softmax (Equation: \ref{eq:softmax}) are not compatible with FHE since they include division although some alternative approximations do exist such as those used by Lee \autocite{lee2021privacy}.
***
\subsubsection{Data Wrangling and Inputs}

Fashion-MNIST is largely pre-wrangled especially if you use one of many forks of the data which present each figure-classification, and image as a one-dimensional feature vector between 0-255 stacked in a CSV file. This means the only two necessary steps toward this data are to normalise between 0-1, and reshape the individual feature vectors back into their original shape of 28x28x1. The feature vector is encrypted and the cyphertext passed in as a signal to node "x" in the sphira network (Figure \ref{fig:fashion_mnist_cg}), and the figure-classification is passed in to node "y" as a separate signal. Whereby our neuronal-firing algorithm (Algorithm \ref{alg:firing}) will propagate these signals thereafter.

\subsubsection{CNN}

\begin{figure}[t!]
  \centering
  \includegraphics[width=\columnwidth]{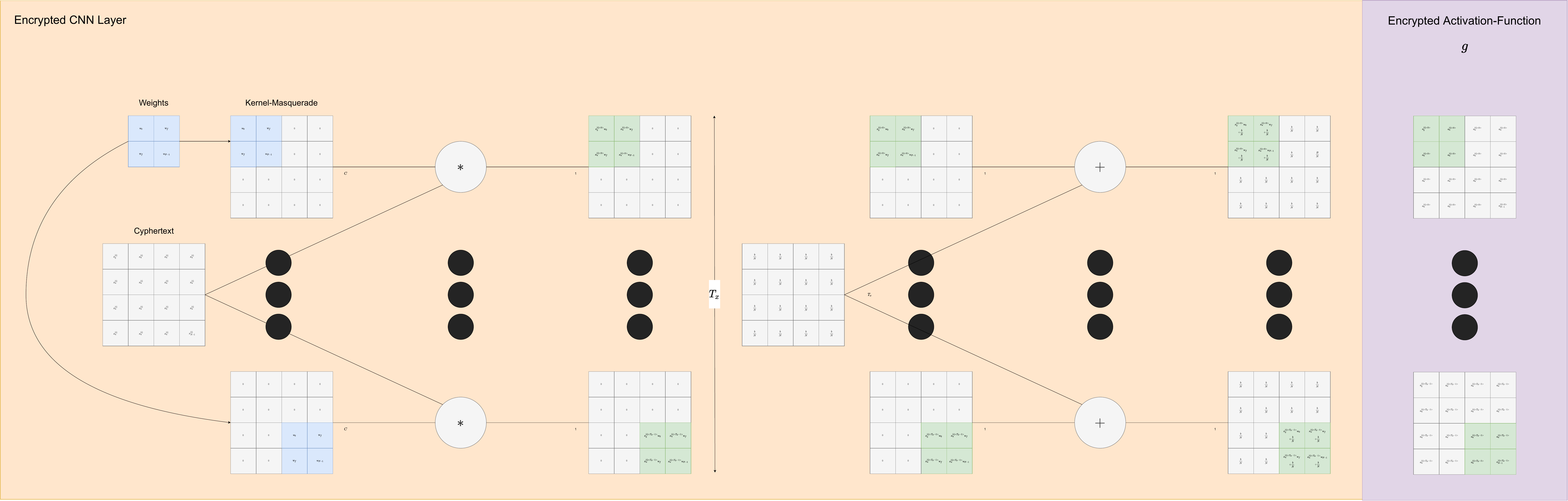}
  \caption{Encrypted convolutional neural network (CNN), this is a particular unusual implementation since there can be no summing of the filters, and instead this sum is commuted in the case where the filter operates on an input that is a single cyphertext (I.E not a composite of multiple cyphertexts). Please see our documentation for closer detail. \autocite{reseal}}
  \label{fig:cg_enc_cnn}
\end{figure}

\begin{equation}
  \label{eq:cnn}
  \begin{aligned}
  a^{(i)<t>} = g(\sum_{t=0}^{T_x-1}(k^{<t>}x^{(i)}+b/N))
  \end{aligned}
\end{equation}

\begin{figure}[t!]
  \centering
  \includegraphics[width=0.7\columnwidth]{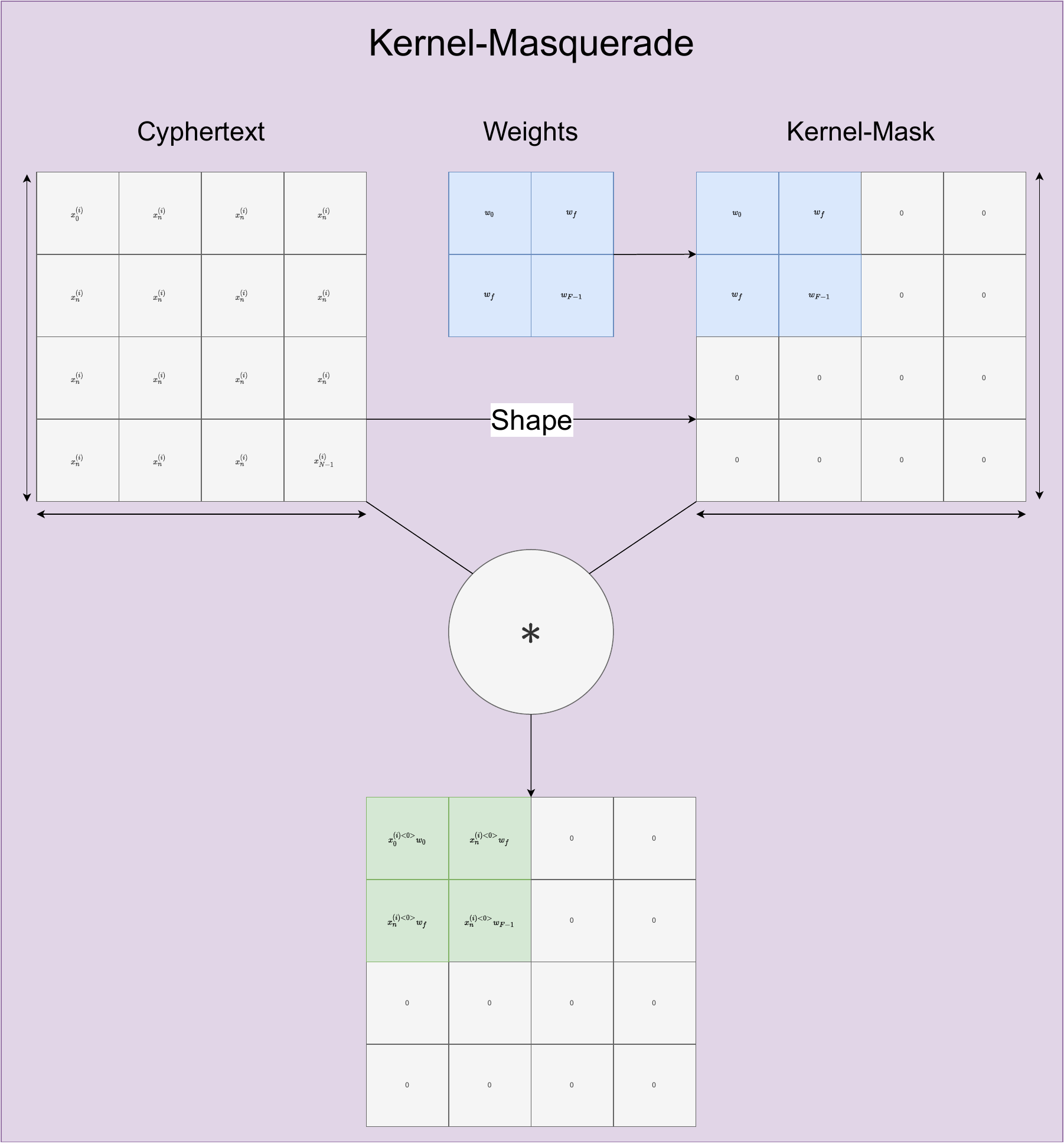}
  \caption{Merged mask and kernel together to create a single sparse kernel which zeros undesired components in the cyphertexts polynomial of values using Hadmard products. Please see our documentation for closer detail. \autocite{reseal}}
  \label{fig:kernel_masquerade}
\end{figure}

As our CNN (yellow in Figure: \ref{fig:fashion_mnist}) we use a biased cross correlation layer (CC) to calculate the product of a given filter against the input cyphertext. We use a SIMO scheme we call kernel masquerading. Here kernel-masquerading shall mean the merging of weights and a respective zeros mask into a sparse n-dimensional array such that they become a single operation conducted on the input cyphertext (Figure: \ref{fig:kernel_masquerade}), reducing the computational depth experienced by the input cyphertext to 1 (multiplication) and allowing for subset operations to be conducted on the cyphertext to selectively pick regions of interest. This is only possible in the plaintext-weights strategy, since this allows the weights to be operated on arbitrarily and selectively to reform them into the shape of the input cyphertext and sparsity of the filter/ kernel. This is a simple operation of which a two and three dimensional variant can be seen in Juvekar's, and Meftah's work \autocite{juvekar2018gazelle, meftah2021doren}. The main drawback of the kernel-masquerade is that if we were to apply a convolutional-kernel-mask on some cyphertext $\varepsilon(x^{(i)})$ we would end up with separate modified cyphertexts $\varepsilon(x^{(i)<t>})$ that correspond to different portions of the data, however we are unable to sum them without a key rotation such that we are summing between different cyphertexts since we cannot fold a cyphertext in on itself. This means we have a choice at this stage, we can either rotate the keys now to reduce complexity or try to save computation time by doing as much processing while the values are encoded in one larger cyphertext which is significantly more efficient from findings in Juvekars work on SISO cyphertexts \autocite{juvekar2018gazelle}.

Since key rotation would make the outputs normally-processable for operations like summation we wont address that variant here instead we choose to see how far we can instead commute this sum to get the maximum performance as far fewer cyphertexts. One thing we can and did do in our CNN implementation is to commute the bias forward to be before summation. so instead of $z = \sum^{N}_{i=0}(x_i w_i) + b$ we decompose $b$ into the product calculation before summation as $z= \sum^{N}_{i=0}(x_i w_i + \frac{b}{N})$ since this is equivalent over the full computation of the cyphertext. We could have simplified to just $z= \sum^{N}_{i=0}(x_i w_i + b)$ if we calculate the gradient with respect to the bias $\frac{df}{db}$ as $\frac{df}{db} = Nx$ instead of $\frac{df}{db} = x$ such that the neural network is effectively aware of this higher contribution of the bias, and it would be naturally accommodated through the gradient descent process.

From here forward special logic/ considerations need to be made to ensure the output cyphertexts of the biased-cross-correlation are treated as a singular un-summed-value. We tried to push this cyphertext through the neural network further but we had to ensure all further operations were both linear and abelian compatible. Take for instance an encoded non-summed sequence as a cyphertext $x$, $x = (1+2+3+4) = 10_{\text{plntxt}}$, then lets try a multiplication $4x = 4(1+2+3+4) = (4*1+4*2 +4*3+4*4) = (4 + 8 + 12 + 16) = 4*10_{\text{plntxt}} = 40_{\text{plntxt}}$, but now lets try a multiplication against itself or another non-summed sequence so for instance a nonlinear $x^2 = (1+2+3+4)(1+2+3+4) = (1*1+1*2+1*3+1*4+2*1+2*2+2*3+2*4+3*1+3*2+3*3+3*4+4*1+4*2+4*3+4*4) = 10^2_{\text{plntxt}} = 100_{\text{plntxt}}$.

This is a problem since we cannot cross-multiply cyphertexts since we cannot select elements from either cyphertext, if we were to attempt to multiply this cyphertext with itself it would calculate the element-wise product of the two. The best we could do if we did want to compute this would be to conduct a key rotation to expose the elements we desired as separate cyphertext but if we are going to do that we would be just as well served by just rotating to sum then passing it through the element-wise product as normal. It is possible to commute the sum further if we use linear approximations of our activation functions like if we take Sigmoid (Equation: \ref{eq:sigmoid}) and its approximation Equation: \ref{eq:sigmoid_approximate} then if we ensure our products will always be between 0-1 through a modified version of batch norm (\autocite{meftah2021doren}) we could then safely use only the linear component of the Sigmoid approximation (Equation: \ref{eq:sigmoid_approximate}) $\sigma(x) \approx \sigma_a(x) = 0.5 + 0.197x$ since it would still closely follow in the -1 to 1 range and loses approximation beyond this range instead of the usual -5 to 5 range the full approximation affords and would also cut down the computational cost. For ourselves we choose to key rotate to encrypt the elements to be summed until such a time as we have fully fleshed out a fully-commuted-sum alternative.

For our cross-correlation activation function we use the more recent ReLU (Equation: \ref{eq:relu}) approximation (Equation: \ref{eq:relu_approximate}) and the derivative of this approximation for backward propagation (Equation: \ref{eq:relu_approximate_derivative}) as its own separate node to allow them to be decoupled and easily swapped out with new or improved variants and so interjection with batch norm is readily variable without having to rewrite existing nodes.


\subsubsection{Dense/ ANN}

\begin{figure}[t!]
  \centering
  \includegraphics[width=\columnwidth]{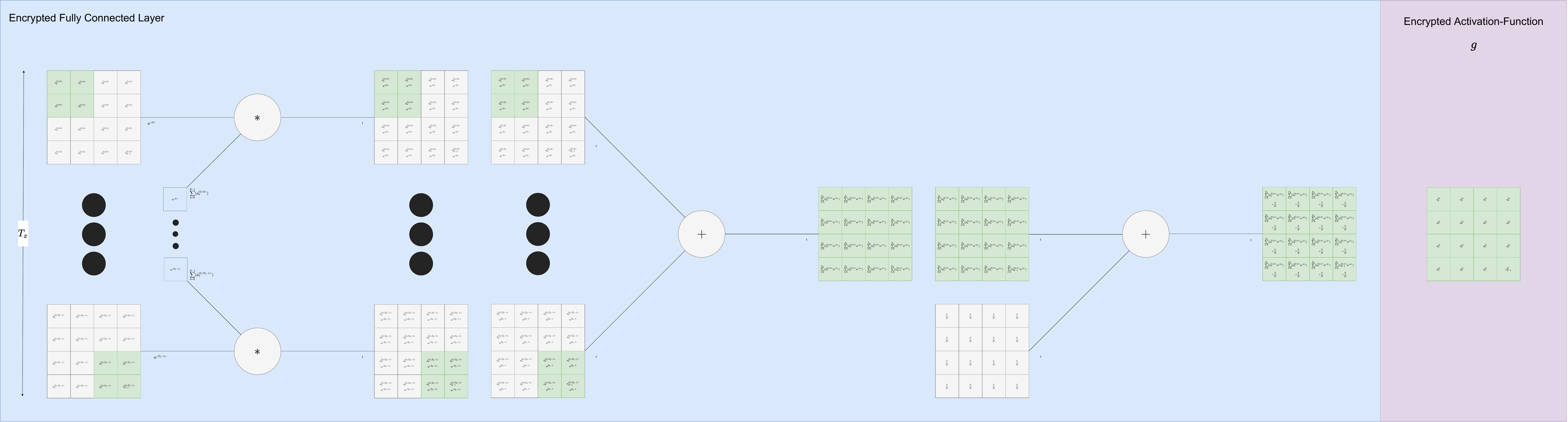}
  \caption{Encrypted variant of an ANN/ dense neural network, usually used in our case to merge divergent times/ branches/ filters back together into a single output. Please see our documentation for closer detail. \autocite{reseal}}
  \label{fig:cg_enc_ann}
\end{figure}

For each of our classes we have a dense fully-connected neuron (Figure: \ref{fig:cg_enc_ann}, red in Figure: \ref{fig:fashion_mnist}) to interpret the activation vector of the biased-cross-correlation and activation combination / CNN. Thus our dense layer is comprised one 10 ANN nodes. There is nothing of any note in this layer other than it must accept multiple cyphertexts that are added together / summed across the first axis, otherwise it behaves almost the same as a standard neuron as depicted in figure. However careful attention should be paid to broadcasting such that the gradient is still correct and we do not attain an exploded result that could fall outside of approximation golden zones like sigmoids -5 to 5.

We accompany each neuron with its own ReLU approximation node before passing the activations on to the different forward evaluation circuits for loss calculation and prediction output.



\subsubsection{Prediction}

Argmax is an effective and quick computation of the highest value in a vector. Since the ANN layer outputs a vector of 10 values one for each class the Argmax function serves to take the highest activation and turn it into a 1-hot-encoded representation of the predicted class. This can be passed into a 1-hot-decoder to attain the predicted class $\hat{y}$. However since argmax relies on the context to find the max, it is necessary to conduct this operation in plaintext on the client side, to effectively pick from this 10 element vector. There is no backpropagation from this branch, it is purely an output branch for providing predictions to the data owner. These stages are pink in Figure: \ref{fig:fashion_mnist_cg}.

\subsubsection{Loss}

The loss calculation stage is represented by purple in Figure: \ref{fig:fashion_mnist_cg}.
Argmax is not an effective function for the purposes of backpropagation of the loss since only one of the ten input ANN neurons would receive all of the gradient multiplied by 1, which does not give the majority of the network much information to update the weights from any single given example. Thus as-per-norm we used a Softmax layer instead which better distributes the gradient between not only what neuron was responsible for positive activation but also the others that should not be activating.

The softmax ensures that all output values summed together equal 1, and that they are effective predicted probabilities of the network that a certain class is what was given in the input. We use a standard categorical cross-entropy (CCE) function to calculate the loss and subsequently the derivative with respect to each of the 10 classes to pass back to the softmax and hence the ANN layer.

The CCE function also receives input/ stimulation from a 1-hot-encoder that encodes the ground-truth $y$ value or the actual class that the input $x$ corresponds to for the purposes of loss calculation.


\begin{equation}
  \label{eq:softmax}
  \begin{aligned}
  \sigma(\overrightarrow{a})_i = \frac{e^{a_i}}{\sum^{K}_{j=1}e^{a_j}}
  \end{aligned}
\end{equation}

It should also be noted that the loss circuit (pink) requires decryption since both softmax and CCE are not FHE compatible operations. There have been proposed ways to allow for softmax to be computed with cyphertexts by Lee \autocite{lee2021privacy} however we were unable to create a working fhe-compatible softmax from what information was available and would require bootstrapping 22 times, and it would still need to be unencrypted for the CCE calculation.
Given this data using the sphira (\ref{fig:fashion_mnist_cg}) network we garnered the following results:


\begin{figure}[t!]
  \centering
  \includegraphics[width=0.8\columnwidth]{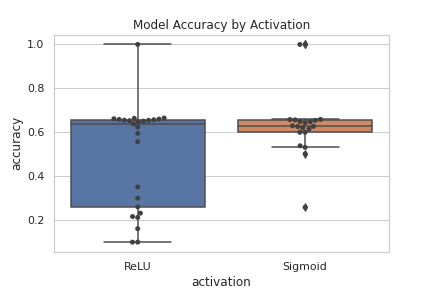}
  \caption{Model performance using different activation functions in the sphira network on the fashion-MNIST dataset. All activations here are their FHE compatible approximations unless otherwise specified. Each dot is a different network, or the same network with a different data type(cyphertext, plaintext). \autocite{reseal}}
  \label{fig:perf_acti}
\end{figure}

\begin{figure}[t!]
  \centering
  \includegraphics[width=0.9\columnwidth]{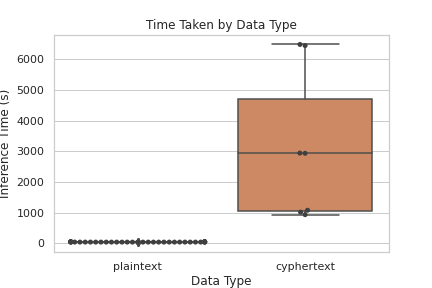}
  \caption{Model inference time, by different types. Plaintext types mean where the graph is run using plaintext data. Cyphertext types mean where the graph is run using cyphertext data. Both plaintext and cyphertext data conforms to the same NumPy API, meaning they can be used interchangeably. Each dot is a different network (I.E differently initialised weights but the same structure), or the same network with a different data type (cyphertext, plaintext). \autocite{reseal}}
  \label{fig:perf_time}
\end{figure}

\subsection{Strawberry Yield Data}

\begin{figure}[t!]
  \centering
  \includegraphics[width=0.8\columnwidth]{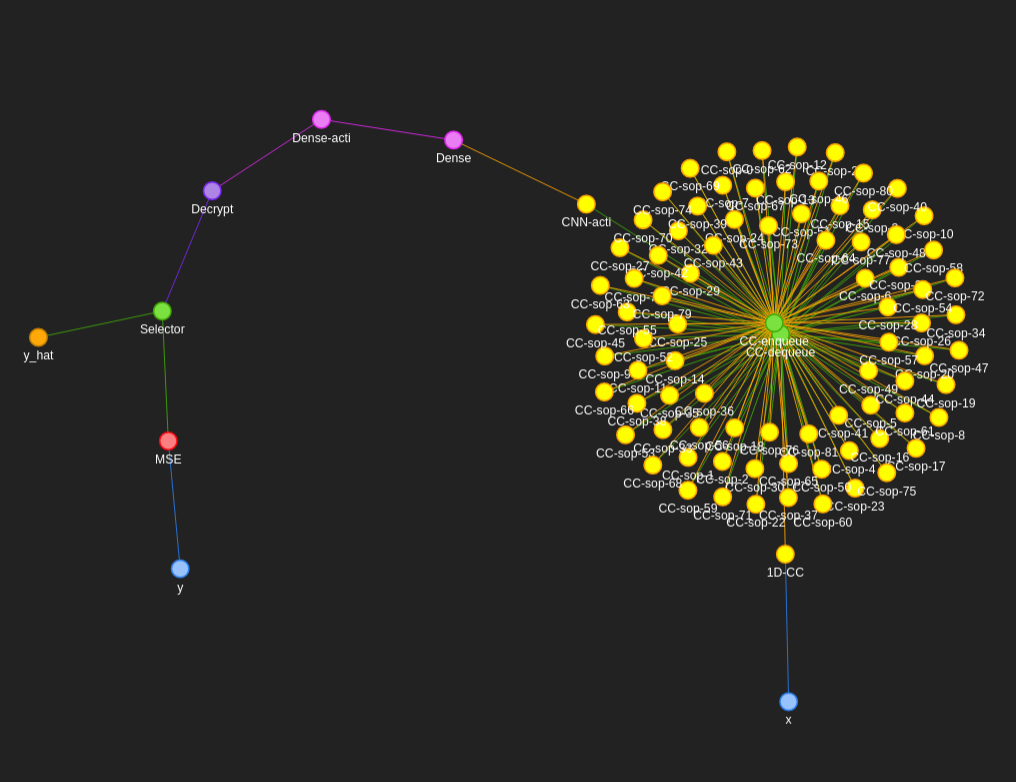}
  \caption{Strawberry yield/ regression computational graph we call "constelation", showing the colour coded graph representation and nodes used to train on strawberry yield, based on environmental factors. Blue are input/ encryption nodes. Yellow are convolutional-related nodes. Green are operational nodes necessary to "glue" the network together. Pink are dense/ ANN nodes. Orange is the output prediction node. Red is the loss calculation node. Purple is an FHE specific node used for decryption of the input data. Please see our documentation for closer detail. \autocite{reseal}}
  \label{fig:constelation}
\end{figure}

Unfortunately we do not have permission to publicise the specific data used in this section. As such we shall preview this application and seek to further elaborate and develop the techniques used here in-depth in future papers. We will only touch briefly here of this data as a means to show how FHE can be used in real world problems effectively, and for a different kind of problem; regression instead of classification.

This data is geared towards yield prediction often weeks in advance. The prediction horizon is typically between one to three weeks ahead of the strawberry fruit becoming ripe. This allows time for logistical constraints such as price negotiation, and picker/ staff scheduling. Thus performance of these predictive models is critically important to ensure all the fruit is being accounted for in negotiations with retailers (and thus can be sold), and that there is sufficient manpower at the point of need to gather this produce. Over predicting can result in insufficient harvests to meet contractual obligations, most likely meaning the yield must be covered by buying other producers yields. However if there is a shortfall of yield from one producer, the factors that lead to that shortfall such as adverse environmental conditions are felt by most other producers in the geographic region. This means that usually (in the UK) the yields must be imported from abroad increasing the price substantially. Conversely if there is under-prediction of yield this results in unsold fruit, which is either sold at a significant discount if possible or destroyed. There is also a clear and significant lack of agriculture data available due to perceived data sensitivity. This affects many forms of agriculture for various reasons. In the soft-fruit industry this tends to be proprietary genetic varieties, and operational specifics such as irrigation nutrition mixtures. However there is a tendency to distrust and a perceived lack of benefit to data sharing, due to no obvious performant outcomes.

Due to a lack of available data, we use historic yield data we gathered in our Riseholme campus polytunnel/ tabletop over two years, and combine this with environmental data experienced by these strawberries leading up to the point-of-prediction. The environmental data includes: wind speed, wind direction, temperature, light-intensity, humidity, precipitation, positions-of-strawberries, yield-per-row-of-strawberries, and many more less significant features. This data also includes irrigation data such as nutrition, soil-moisture, soil-temperature, irrigation-status. We normalised, one-hot encoded categorical variables and split the data (80-20) randomly into training and test sets. We then further subdivided the training set into validation sets for model selection purposes.

We applied a 1D/ time-series CNN (Equations: \ref{eq:cnn}, \ref{eq:sigmoid_approximate} as depicted in Figure: \ref{fig:cg_enc_cnn}) followed by a dense ANN, and Sigmoid again as depicted in Figure: \ref{fig:cg_enc_ann}) to summarise the feature vector and to predict the output/ yield of the strawberries. This prediction will then be based on the environment the strawberries experienced leading up to the point of prediction. Given our data, and our feature engineering, we were able to obtain the following outcomes in Table: \ref{tab:strawberry}.

\begin{table}[t!]
\begin{tabular}{ll}
Days Ahead & \begin{tabular}[c]{@{}l@{}}Mean Absolute Percentage Error\\ (MAPE)\end{tabular} \\ \hline
7          & 8.001                                                                               \\
14         & 14.669                                                                              \\
21         & 22.326
\end{tabular}
\caption{Table of predictive results of the constellation network (Figure: \ref{fig:constelation}) predicting strawberry yield.}
\label{tab:strawberry}
\end{table}

%

\subsection{Equations} \label{equations}

\subsubsection{Sigmoid}

\begin{equation}
  \label{eq:sigmoid}
  \begin{aligned}
    \sigma(x)=\frac{1}{1+e^{-x}}
  \end{aligned}
\end{equation}

\subsubsection{ReLU}

\begin{equation}
  \label{eq:relu}
  \begin{aligned}
    R(x)=\text{max}(0,x)
  \end{aligned}
\end{equation}

\subsubsection{Sigmoid-Approximate}

\begin{equation}
  \label{eq:sigmoid_approximate}
  \begin{aligned}
    \sigma(x) \approx \sigma_a(x) = 0.5 + 0.197x + -0.004x^3 
  \end{aligned}
\end{equation}

\subsubsection{ReLU-Approximate}

\begin{equation}
  \label{eq:relu_approximate}
  \begin{aligned}
    R(x) \approx R_a(x) = \frac{4}{3\pi q}x^2 + \frac{1}{2}x + \frac{q}{3\pi}
  \end{aligned}
\end{equation}

\subsubsection{Sigmoid-Derivative}

\begin{equation}
  \begin{aligned}
    \frac{d\sigma(x)}{dx} = \frac{e^{-x}}{(1+e^{-x})^2} = (\frac{1+e^{-x}-1}{1+e^{-x}})(\frac{1}{1+e^{-x}}) = (1-\sigma(x))\sigma(x)
  \end{aligned}
\end{equation}

\subsubsection{ReLU-Derivative}

\begin{equation}
  \begin{aligned}
    \frac{dR(x)}{dx} = \begin{cases} 1, & \text{if}\ x>0 \\ 0, & \text{otherwise} \end{cases}
  \end{aligned}
\end{equation}

\subsubsection{Sigmoid-Approximate-Derivative}

\begin{equation}
  \begin{aligned}
    \frac{d\sigma(x)}{dx} \approx \frac{d\sigma_a(x)}{dx} = 0.197 + -0.012x^2
  \end{aligned}
\end{equation}

\subsubsection{ReLU-Approximate-Derivative}

\begin{equation}
  \label{eq:relu_approximate_derivative}
  \begin{aligned}
      \frac{dR(x)}{dx} \approx \frac{dR_a(x)}{dx} = \frac{8}{3\pi q}x + \frac{1}{2}
  \end{aligned}
\end{equation}


\section{Results}
As can be seen in Figure: \ref{fig:perf_acti} using the same network sphira (Figure: \ref{fig:fashion_mnist_cg}) with different approximated activation functions Sigmoid (Equation: \ref{eq:sigmoid_approximate}) and ReLU (Equation: \ref{eq:relu_approximate}) dramatically effects the precision of the neural network over multiple training attempts with randomised weights. However the accuracy of both on average is roughly equal within a few percent. This shows that in our implementation at least that this ReLU approximation being backpropagated may indeed cause some instability, more frequently. Sigmoid in contrast is a static approximation which may be part of the reason for its greater stability, and thus consistency provided randomised weights to the rest of the network.
We can see that both the sphira (Figure: \ref{fig:fashion_mnist_cg}) and constellation (Figure \ref{fig:constelation}) networks can produce acceptable results on the testing set while computing over cyphertexts and plaintexts. Our networks can be seen working in both classification and regression problems, Fashion-MNIST, and strawberry yield prediction respectively.

We find however that in our strawberry yield prediction one of the weaknesses of our approach was to completely randomise the sequences, as some sequences could possibly overlap, meaning that the network may have at least some prior experience of the gap between point-of-prediction and point-predicted. This should be an area of possible future expansion is to split the data differently by time, and use only environmental data in the future that is distinct.
Another area where improvement could be garnered is by using smaller but bootstrappable cyphertexts, this may reduce predictive performance of the networks since each bootstrapping operation would incur a noise penalty, but this would significantly improve the speed of computation since we could use smaller cyphertexts that take less time to transverse, transmit, and compute. We can see from Figure: \ref{fig:perf_time} that the time taken for computing plaintexts is relatively small, producing results rapidly. The same network however provided cyphertexts computes near equivalent results, but significantly slower. Not only are cyphertexts more time intensive, but they are also significantly more space intensive. We could see during cyphertext inference anywhere from 72-80GiB of RAM usage, meaning this is certainly not plausible on low-specification machines. We could see significant gains in computational performance if we added more rotation nodes to refresh the cyphertext more frequently, to limit the number of levels it would contain, and thus the size of the cyphertext. However in our case we wanted to reduce the number of rotations as in practical applications this would result in more transmissions from client-server which itself can be an expensive operation. This is a good example of why bootstrapping while incurring a high cost itself, could save computational time and space in the future, when it is more widely available.

The absolute performance of the two models in Figure \ref{fig:perf_acti} and Table \ref{tab:strawberry} is acceptable despite being fairly shallow models compared to those used in many normal deep learning models. Our absolute performance is probably quite limited by the shallowness of these models, in that the model may not be complex enough to properly model some of these problems. In particular there are a plethora of standard models that achieve 90\% accuracy or greater, many of which use 2-3 convolutional layers, with batch-normalisation, and max-pooling. Clearly we cannot ordain context from a cyphertext making max-pooling impossible however we can and do use strides as a way to reduce the dimensionality in a similar way that max-pooling does. There have also been proposals for batch-normalisation that involve multiplying by small fractions that are occasionally recalculated, however this is quite complex and not something we have been able to implement ourselves as of yet. This would however stabilise the activations between nodes, and reduce the likelihood of escaping the dynamically predicted range in the ReLU approximation causing the in-precision in Figure \ref{fig:perf_acti}.

Given that we can get acceptable performance, in different scenarios like agricultural yield regression and image classification, this opens avenues for data sharing. There are two avenues in particular, through encryption, and through trust. In our case we assume a semi-honest threat model, yet we have outlined a way of computation that does not need to reveal any data to the third party. This means if we can provide sufficient predictive performance then there are few barriers preventing sharing of encrypted data for inference. There is of course the notable exception of any yet unknown vulnerabilities in the underlying FHE scheme with default parameters provided by MS-SEAL. The other avenue of data sharing that FHE fosters is that of trust. Given a track record of reliable data processing in the encrypted form, that this could lead to an increased awareness of the gains of deep learning applied to various fields. With this greater awareness, and track record, it could be surmised that it is more likely that over time the data owners might choose to share data in the perceived-sensitivity scenario.

\section{Discussion and Limitations}
Considering the importance around ascertaining privacy when developing new machine learning methodologies, it is paramount that we start scaling up research on privacy-enabled machine learning. This should take place in tandem with showing how real-life problems, e.g. strawberry yield forecasting, can be tackled with such methodologies, which is what this paper has contributed to, results-wise as well (Table \ref{tab:strawberry}). Nevertheless, our work on encrypted deep learning has certain limitations we would like to highlight:
    
    \textbf{FHE Training}; In this paper we laboriously implement, describe, and show how encrypted deep learning inference can be conducted. However there is little reference to encrypted learning, that is to say where a neural network is trained on cyphertexts. This is due to multiple limitations prevalent in the field such as the lack of FHE compatibility with certain functions, such as loss functions. This is an active area of research which we and the broader research community are actively working on, to complete the encrypted learning-to-inference chain. Another issue with cyphertext training for example is when do we decide to stop training? As we cannot see the results, and thus cannot gauge whether the network has under or over-fit. This is a particularly interesting and challenging problem which we seek to also tackle in future. Here however it can be thought that the models would be pre-trained or transferred from a similar problem to avoid privacy leaks involved in training.
    
    
    \textbf{LFHE}; As previously mentioned our work here is over Levelled-FHE, where we create optimised circuits for cyphertexts with discrete scales and primes, which we swap down for each multiplication, a "level". LFHE is FHE without bootstrapping. Bootstrapping is an expensive operation that refreshes the cyphertexts levels allowing for an effective limitless depth to computations (albeit with noise), while also helping to keep cyphertexts smaller than their LFHE counterparts. Smaller cyphertexts can be operated on faster, but bootstrapping in small circuits can often outweigh the benefit of using a smaller but bootstrapped cyphertext, due to how expensive of an operation it is. This limitation comes from a lack of bootstrapping support in MS-SEAL. Once bootstrapping is supported however existing networks we propose here will still be compatible assuming appropriate NumPy-container abstractions of FHE will be passed in.
    
    \textbf{State-of-the-Art Neural Networks}; While this work particularly focuses on ANN and CNN neural networks, these are not current State-of-the-Art networks for many tasks. In particular in future we intend to continue to work applying FHE to existing networks such as transformers which are SotA in sequence tasks. However much work remains in mimicking certain functions of transformers in an FHE compatible manner \autocite{falcetta2022privacy}. We also believe while we could compute privately, we can significantly improve the performance of the predictions themselves with more performant network architectures like transformers. We could then draw more comprehensive comparisons between encrypted and unencrypted deep learning for yield forecasting and other applications.

\section{Conclusion and Future Work}
In this paper, we have shown how FHE can be automatically parameterised directly from multi-directed graphs for neural networks, using groups and a variation of the travelling salesman problem for costs. It was demonstrated how multi-directed graphs can be used in an FHE compatible manner with FHE compatible nodes to facilitate encrypted deep learning. We have also evaluated a recent ReLU approximation (with additionally backpropagated approximation range), against the Sigmoid activation function, finding it slightly less accurate but much less precise due to instabilities in the weight initialisation. The proposed encrypted deep learning procedures  were  utilised in both classification and regression problems. For the former, we used an open dataset Fashion-MNIST with open-source reproducible code examples to aid reproduction and experimentation. For the latter, We demonstrated how our methods can be used in an real world (sensitive) problem predicting strawberry yield, paving the way to introduce such a technology at scale in the agri-food sector. We believe that our implementation is the most comprehensive encrypted deep learning library currently available, now with automatic FHE parameterisation, traversal, cross-compatible/ interoperable NumPy custom-containers, documentation and expandability for future distributed or GPU accelerated computations with FHE, using the state-of-the-art RNS-CKKS FHE scheme provided by the MS-SEAL back end.

However, there is still much research that needs to be conducted, in particular with FHE and training. Encrypted deep learning is not a solution currently to any problem that relies on very specific data that is very dissimilar to other problems, meaning we cannot transfer some understanding in a private manner. We are still limited by multi/parallel-processing, however in the case of Python-FHEz we leave the back end open-ended following the NumPy custom container specification such that this gap can be easily retrofitted later, just like Dask and CuPy have for standard NumPy.

Finally with encrypted deep learning we can open avenues for data sharing that have previously been untenable in the face of their rightful privacy concerns. The more of the pitfalls of FHE that are solved, and the more usable encrypted deep learning becomes, the more likely we are to see it provide some critical predictive service to improve fields like agriculture, and medicine.

\ifCLASSOPTIONcompsoc
  \section*{Acknowledgments}
\else
  \section*{Acknowledgment}
\fi

The authors would like to thank the British Biotechnology and Biological Sciences Research Council (BBSRC) in collaboration with Berry Gardens Growers and the University of Lincoln for funding and support.

\ifCLASSOPTIONcaptionsoff
  \newpage
\fi



%
%
%
\printbibliography

\end{document}